\documentclass[letterpaper, 10 pt, conference]{ieeeconf}  

\IEEEoverridecommandlockouts                              

\usepackage{amsmath,amsfonts}
\usepackage{algorithmic}
\usepackage{algorithm}
\usepackage{array}
\usepackage[caption=false,font=normalsize,labelfont=sf,textfont=sf]{subfig}
\usepackage{multirow}

\usepackage{textcomp}
\usepackage{stfloats}
\usepackage{url}

\usepackage[colorlinks,urlcolor=blue,linkcolor=blue,citecolor=blue]{hyperref}
\usepackage{verbatim}
\usepackage{graphicx}
\usepackage{cite}
\usepackage[flushleft]{threeparttable}
\usepackage{makecell}

\usepackage{xcolor}

\title{\LARGE \bf
A Case Study on Visual-Audio-Tactile Cross-Modal Retrieval}

\author{Jagoda Wojcik$^{*}$, Jiaqi Jiang$^{*}$, Jiacheng Wu and Shan Luo 
\thanks{This work was supported by the EPSRC project ``ViTac: Visual-Tactile Synergy for Handling Flexible Materials" (EP/T033517/2).}
\thanks{ All the authors are with Department of Engineering, King's College London, London WC2R 2LS, U.K. Emails: {\tt\small \{jagoda.wojcik, jiaqi.1.jiang, jiacheng.2.wu, shan.luo\}@kcl.ac.uk }}
\thanks{$*$ represents equal contributions.}
}

\begin{document}

\maketitle
\thispagestyle{empty}
\pagestyle{empty}

\begin{abstract}
Cross-Modal Retrieval (CMR), which retrieves relevant items from one modality (e.g., audio) given a query in another modality (e.g., visual), has undergone significant advancements in recent years. This capability is crucial for robots to integrate and interpret information across diverse sensory inputs. 
However, the retrieval space in existing robotic CMR approaches often consists of only one modality, which limits the robot's performance.
In this paper, we propose a novel CMR model that incorporates three different modalities, i.e., visual, audio and tactile, for enhanced multi-modal object retrieval, named as \textit{VAT-CMR}.  
In this model, multi-modal representations are first fused to provide a holistic view of object features.
To mitigate the semantic gaps between representations of different modalities, a dominant modality is then selected during the classification training phase to improve the distinctiveness of the representations, so as to improve the retrieval performance.
To evaluate our proposed approach, we conducted a case study and the results demonstrate that our VAT-CMR model surpasses competing approaches. Further, our proposed dominant modality selection significantly enhances cross-retrieval accuracy.

\end{abstract}

\section{INTRODUCTION}
In recent years, Cross-Modal Retrieval (CMR), i.e., querying data in one modality (e.g., audio), to retrieve relevant items from another modality (e.g., vision) has made significant strides~\cite{kaur2021comparative,luo2018vitac,luo2017robotic}. This progress has been driven by the exponential growth of multi-modal data, in various forms of images, texts and audio, on the internet and in our daily lives. CMR holds great promise for applications such as healthcare, where it could align medical imaging with related patient profiles, thereby improving diagnostic accuracy. In the context of robotics, CMR enables the processing and interpretation of information across diverse sensory inputs, such as vision and touch, empowering robots to adapt and interact more effectively with their environment~\cite{lee2019touching, 10160373,luo2015localizing}.

\begin{figure}[t]
  \includegraphics[width=\linewidth]{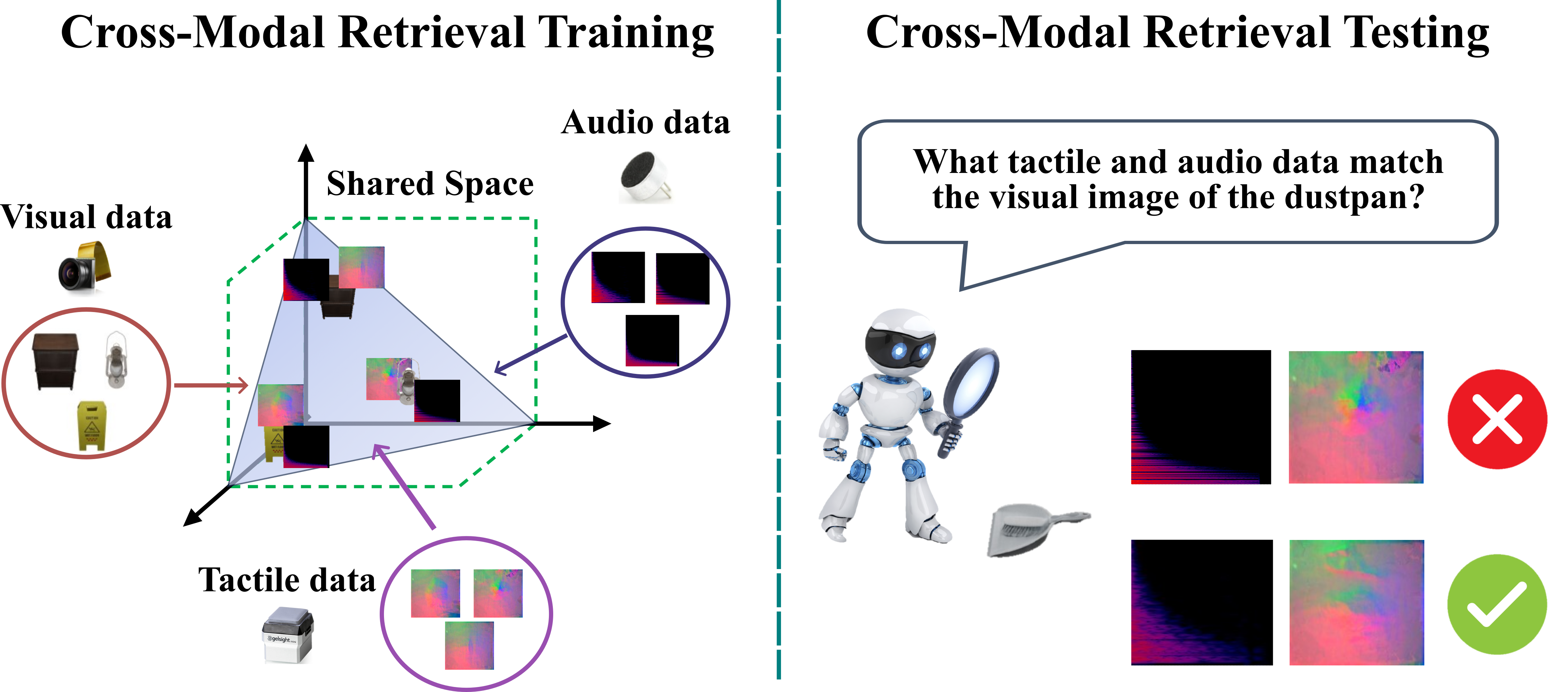}
   \caption{Illustration of visual-audio-tactile cross-modal retrieval. Left: The visual, audio and tactile representations of the same object converge within a shared space. Right: the robot retrieves the corresponding audio and tactile data when provided with a visual image of the dustpan. }
\label{fig:demo}
\end{figure}

Many existing CMR approaches within the field of robotics are limited by their focus on single-modality retrieval and reliance on bi-modal CMR networks~\cite{pecyna2022visual,jiang2022shall}. While relying on a single-modality may compromise the retrieval accuracy due to restricted information scope, the use of multiple bi-modal CMR models in dealing with more than two modalities~\cite{gao2022objectfolder} increases the computational complexity and reduces the overall efficiency.

In contrast, human perception seamlessly integrates information across multiple modalities, such as vision, sound, and touch, to form a cohesive understanding of the environment. This natural ability to cross-reference sensory information leads to more robust and accurate information retrieval. Research on human sensory perception ~\cite{ghazanfar2006neocortex,stein2008multisensory} has demonstrated that the integration of multiple modalities can facilitate the development of more effective neural representations, thereby enhancing cognitive performance. 


In this work, we propose a novel CMR model that incorporates three distinct modalities for object retrieval, named \textit{VAT-CMR}, as illustrated in Fig.~\ref{fig:demo}. In the model, multi-modal representations are first learned to offer a holistic representation of the object features. This approach enhances the disambiguation of latent spaces that might otherwise remain ambiguous when confined to a single modality. To improve the alignment among the diverse modalities, we use an attention mechanism during the multi-modal feature fusion stage.
Additionally, we introduce the concept of \textit{dominant modality selection} during the classification training phase. This approach differs from conventional methods, which often optimise based on a concatenated representation of features from multiple modalities. The emphasis on a dominant modality aims to effectively mitigate the semantic gap between modalities, thereby boosting the performance of our retrieval model.


To evaluate our
proposed VAT-CMR approach, we conduct a case study and collect a synthetic dataset featuring 20 objects with data from three modalities, i.e., vision, audio, and touch. Our experiments demonstrate that VAT-CMR outperforms the state-of-the-art CMR methods, showcasing a noteworthy improvement in Mean Average Precision (MAP) when the query modality is either vision or touch. Additionally, through an ablation study, we find that both the attention feature modules and the incorporation of dominant modality selection contribute to enhanced retrieval accuracy by 0.04 and 0.05, respectively.

    
    

Our contributions can be summarised as follows:
\begin{itemize}
    \item We propose VAT-CMR, a novel CMR model that utilises multi-modal feature representations for retrieval tasks;
    
    \item We introduce the concept of dominant modality-based training for CMR, which enhances the retrieval performance;
    
    \item The proposed VAT-CMR outperforms the state-of-the-art approaches, with code publicly available\footnote{ \url{https://github.com/jagodawojcik/VAT-CMR}}.
\end{itemize}

The rest of the paper is structured as follows: Section~\ref{Section:2} provides an overview of related works; Section~\ref{Section:3} introduces our VAT-CMR framework; Section~\ref{Section:4} details our synthetic dataset and the evaluation metrics used; Section~\ref{Section:5} analyses the experimental results. Finally, Section~\ref{Section:6} presents the discussion and summarises the work.

\section{Related Works}
\label{Section:2}
The exploration of Cross-Modal Retrieval (CMR) has gained increasing attention in recent years, due to the rapid proliferation of multi-modal data in robotics, such as images, text, audio and tactile readings. Existing CMR methods fall into two categories: traditional multi-modal representation learning methods and deep multi-modal representation learning methods.

Early techniques for CMR relied on simplistic representations, with Canonical Correlation Analysis (CCA)~\cite{rasiwasia2010new} being one of the popular methods. CCA maximises the correlation between modalities, employing a semantic space to measure similarity. Other methods, like Partial Least Squares (PLS)~\cite{rosipal2005overview} and Bilinear Model (BLM)~\cite{sharma2012generalized}, also aimed at learning latent common spaces but often faced scalability and generalisation challenges.

Recent advancements in CMR are shaped by deep network-based representation learning~\cite{srivastava2012multimodal}. 
Notable examples include Deep Canonical Correlation Analysis (DCCA)~\cite{andrew2013deep}, which learns intricate nonlinear transformations to ensure a strong linear correlation between bi-modal data representations. Deep Canonically Correlated Autoencoders (DCCAE)~\cite{wang2015deep} extend these concepts through reconstruction objectives. In the realm of adversarial learning, Adversarial Cross-Modal Retrieval (ACMR)~\cite{wang2017adversarial} employs a feature projector, a modality classifier and a triplet constraint to establish an effective common subspace. 
Deep Supervised Cross-modal Retrieval (DSCMR)~\cite{zhen2019deep} focuses on 
minimising discrimination loss in both the label space and the common representation space to learn discriminative features. 


Recently, there have been works on cross-modal retrieval among visual, audio and tactile modalities~\cite{liu2018active, zheng2019cross}. Liu et al.~\cite{liu2018active} investigate active visual-tactile cross-modal matching using a dictionary learning model, while their work~\cite{liu2018surface} introduces a framework for weakly paired fusion of tactile and auditory modalities and cross-modal transfer for the visual modality. Zheng et al. propose a novel Discriminant Adversarial Learning (DAL) method, addressing intra-modal discrimination and inter-modal consistency for visual-tactile cross-modal retrieval in a unified training process. Despite these advancements, there is a gap in exploring multi-modal retrieval representations for cross-modal retrieval with multiple modalities.


\begin{figure*}[t]
  \includegraphics[trim=0 0 0 -10, clip, width=\linewidth]{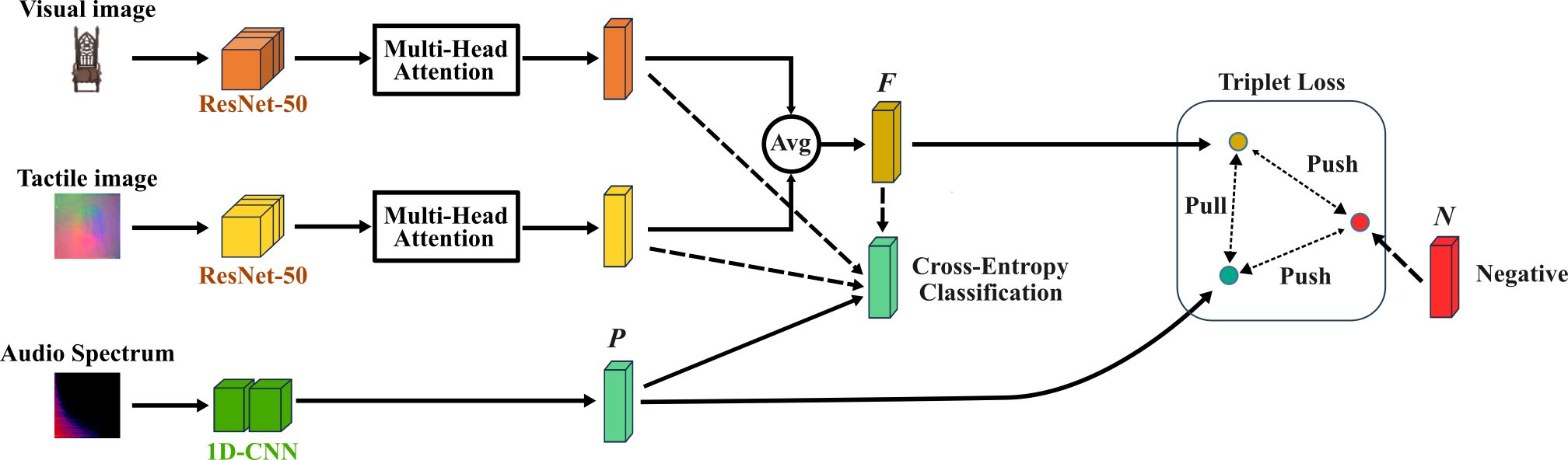}
   \caption{\textbf{Overview of our VAT-CMR model (with audio as the query modality for example).} From left to right: First, VAT-CMR takes a visual image, a tactile image and an audio sample as input. These inputs are processed through three separate neural network branches. Multi-head attention modules are used to fuse the feature representations from the two retrieval modalities. With the fused feature representation $F$, and the positive input modality feature $P$, the model is trained using cross-entropy loss where audio serves as the dominant modality and is directly linked via a solid line. The dashed lines connected to the cross-entropy module represent the cases when other modalities are selected as the dominant modality. Finally, a triplet loss function is employed to map the features extracted from the last hidden layer of each branch to a cross-sensory embedding space. $N$ on the right side represents a negative sample used in triplet loss training.}
\label{fig: overview}
\end{figure*}

\section{Methodology}
\label{Section:3}
In this section, we first outline the problem formulation and then introduce our proposed Vision-Audio-Touch Cross-Modal Retrieval (VAT-CMR) model, with an overview of the model illustrated in Fig.~\ref{fig: overview}. Finally, we provide details on the training methodology to facilitate the reproduction of this work. 

\subsection{Problem Formulation}
The Cross-Modal Retrieval (CMR) problem addressed in this work considers three types of data: visual, audio and tactile readings. Let $v \in \mathbb{R}^{d_v}$, $a \in \mathbb{R}^{d_a}$, and $t \in \mathbb{R}^{d_t}$ represent the visual features, audio features, and tactile features derived from visual images, audio data and tactile data, respectively. Here, $d_v$, $d_a$ and $d_t$ denote the dimensions of the latent representations of the vision, audio, and touch modalities, respectively. Each sample of visual, audio and tactile data is associated with a corresponding one-hot category label $y_i \in \mathbb{R}^{C}$, where $C$ represents the number of categories in the dataset.

Given a set of training samples  $V = \left\{v_1, v_2, \ldots, v_n \right\}$, $A = \left\{a_1, a_2, \ldots, a_n \right\}$ and $T = \left\{t_1, t_2, \ldots, t_n \right\}$ extracted from paired visual, audio, and tactile data, VAT-CMR first employs supervised learning to establish a shared semantic space. The primary objective is to ensure close alignment among the corresponding visual, audio, and tactile representations, facilitating instance retrieval across modalities when provided with data from a specific modality. This alignment is achieved through a function $f(\cdot)$ that maps features from all modalities into a common semantic space. The training goal is to maximise the semantic similarity between multi-modal representations of the same object. Euclidean distance is employed as a similarity metric $sim(\cdot,\cdot)$, which is used to evaluate the distance between query and retrieval feature representations:
\begin{equation}
f: V \times A \times T \rightarrow S
\end{equation}
where $S$ represents the shared semantic space.

\subsection{The proposed VAT-CMR model}

To address the visual-audio-tactile cross-modal retrieval problem, we develop a three-branch network. As shown in Fig.~\ref{fig: overview}, our VAT-CMR model takes a visual RGB image, a tactile image from an optical tactile sensor~\cite{yuan2017gelsight,lambeta2020digit,gomes2020geltip,cao2023touchroller} and an audio sample as input. Within each branch, the three distinct modalities are processed through disjoint networks during the initial layers to capture the modality-specific features. An attention mechanism is used to fuse the multi-modal representations of the retrieval modalities. Finally, a cross-entropy loss based on the selected dominant modality is employed to obtain final feature representations. These are then mapped to the common latent space using triplet loss.

\subsubsection{Disjoint neural networks architecture} Firstly, we extract the features from visual images, audio patches, and tactile images to represent the object from different modalities. Specifically, two pre-trained ResNet50 models~\cite{he2016deep} are fine-tuned using the visual images and the tactile images generated for selected objects. The visual features $v$ and tactile features $t$ are extracted from the last hidden layer of the fine-tuned models. 
To obtain audio feature embeddings, we use a 1D-CNN architecture, which comprises three convolutional layers followed by a pooling layer to reduce the spatial dimensions of the feature maps and standardise the size of the output representations. Subsequently, the pooling layer's output is fed into a fully connected layer with the intended output dimension $a$.  

\subsubsection{Attention mechanism} The multi-head attention mechanism enables the model to concurrently process information from different representation subspaces at various positions. Following the approach introduced in Vaswani et al.~\cite{vaswani2017attention}, we utilise a multi-head attention mechanism to highlight shared features across two different modalities. In contrast to a single attention head, where critical information could be diluted through averaging, the multi-head mechanism maintains the integrity of these features.
The fused representation $F$ that combines two streams of modalities can be given by:
\begin{equation}
\begin{aligned}
    F = \frac{1}{2} \left( \operatorname{MultiHead}(Q_t, K_v, V_v) + \operatorname{MultiHead}(Q_v, K_t, V_t) \right),
\end{aligned}
\end{equation}
where $\operatorname{MultiHead(\cdot,\cdot,\cdot)}$ is the multi-head attention mechanism~\cite{vaswani2017attention}; $v$ and $t$ in $(Q_t, K_v, V_v)$ and $(Q_v, K_t, V_t)$ signify the visual and tactile modalities, respectively. Here, we present the equation with audio as the query modility for example, and with visual and tactile modalities fused. However, within the VAT-CMR framework, any combination of these three modalities can be selected for integration and we conducted experiments across all possible combinations of these modalities.




\subsubsection{Cross-entropy loss for dominant modality optimisation} 

In previous studies, a prevailing approach in guiding model training across multiple modalities has been to employ a common feature vector, as highlighted by~\cite{zheng2020lifelong, gao2022objectfolder}. However, handling multiple modalities simultaneously can increase the complexity of the learning task, potentially leading to degraded integrated representations due to noisy data or less discriminative features in certain modalities, ultimately resulting in suboptimal performance. Our series of experiments reveals that a more effective strategy for multi-modal representation learning involves focusing the learning process on a selected modality. By empirically evaluating the contributions of each modality to the learning process, we identify the most beneficial modality, which is taken as the dominant modality, as the guiding force for optimisation. By focusing on a single modality, we reduce the complexity of the learning process, making it easier for the model to learn meaningful representations. In this way, instead of utilising the combined representations of multiple modalities, we use the last hidden layer of the disjoint neural network pathway for the dominant modality to compute the cross-entropy loss. As a result, cross-entropy loss can be obtained by:
\begin{equation} 
   L_{dominant}(p, q) = -\sum_{i=1}^{K} p(x_i) \log(q_{dominant}(x_i))
\end{equation}
where \(p(x_i)\) represents the true probability distribution for the target class \(x_i\), while \(q_{{dominant}}(x_i)\) denotes the predicted probability distribution from the selected dominant modality for class \(x_i\). The loss calculation is conducted at the last hidden layer of disjoint pathways, and encompasses all \(K\) object classes. This ensures a thorough evaluation of the model's classification performance, focusing on the modality deemed most informative.


Focusing on a single modality during training offers significant advantages for classification tasks, enhancing the model's ability to discern relevant features. Importantly, this approach also yields long-term benefits for cross-modal retrieval processes. By emphasising one modality, the model becomes more adept at extracting information, leading to improved retrieval accuracy across modalities. This versatility enhances the model's utility and effectiveness in various tasks, making it a valuable tool in multi-modal applications.

\subsubsection{Cross-modal correlation learning}
To achieve the goal of enhancing the similarity between multi-modal representations of the same object, in this work we utilise the triplet loss, as shown in Fig.~\ref{fig: overview}. The Euclidean distance, also known as the L2-norm, is used to calculate the similarity between different representations. To this end, the triplet loss function can be computed as follows:
\begin{equation}
\begin{aligned}
L(F,P,N) = max(||F - P||^2 
- ||F - N||^2 + \alpha, 0)
\end{aligned}
\label{eq: triplet-loss}
\end{equation}
where $F$, $P$ and $N$ represent the fused representations based on the attention module, the positive input modality of the same class, and negative input which in this context denotes a foreign object, respectively. The parameter $\alpha$ is a safety margin, ensuring that the model does not trivialise the equation to zero by equalising the three embeddings.

\subsection{Training Details}
To enable easy replication of our work, we outline the training details in this subsection.
We employed a batch size of 5 for all cross-entropy training tasks due to extensive memory requirements associated with larger batch sizes. All cross-entropy training tasks span 50 epochs. Adam was chosen as the gradient optimisation algorithm, with learning rates set at 0.001 and 0.0001 for the cross-entropy and triplet loss training stages, respectively. One critical hyperparameter, the triplet loss margin, was empirically determined to be 0.5 after extensive testing.


\begin{figure*}[t]
\centering
  \includegraphics[trim=0 0 0 -5, clip, width=\linewidth]{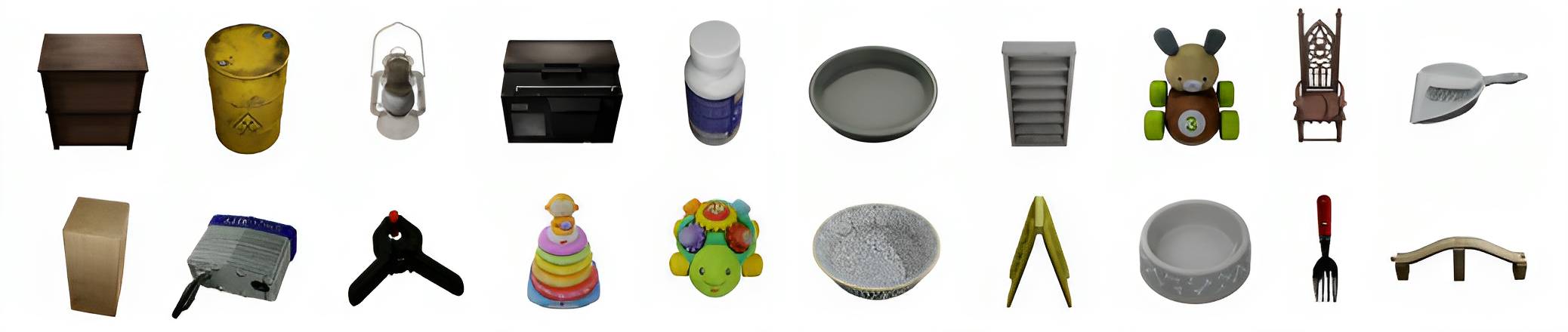}
   \caption{\textbf{Objects used in our experiments.} In total, there are 20 objects in our experiments, taken from the ObjectFolder dataset~\cite{gao2022objectfolder2}, each with unique geometric characteristics and made from materials with distinct properties.}
\label{fig: object}
\end{figure*}

\section{Experimental Setup}
\label{Section:4}
In this section, we first introduce the process of synthetic dataset generation. Then we introduce the evaluation metrics for cross-model retrieval. 

\begin{figure}[t]
   \centering
   \includegraphics[trim=0 0 0 -5, clip, width=\linewidth]{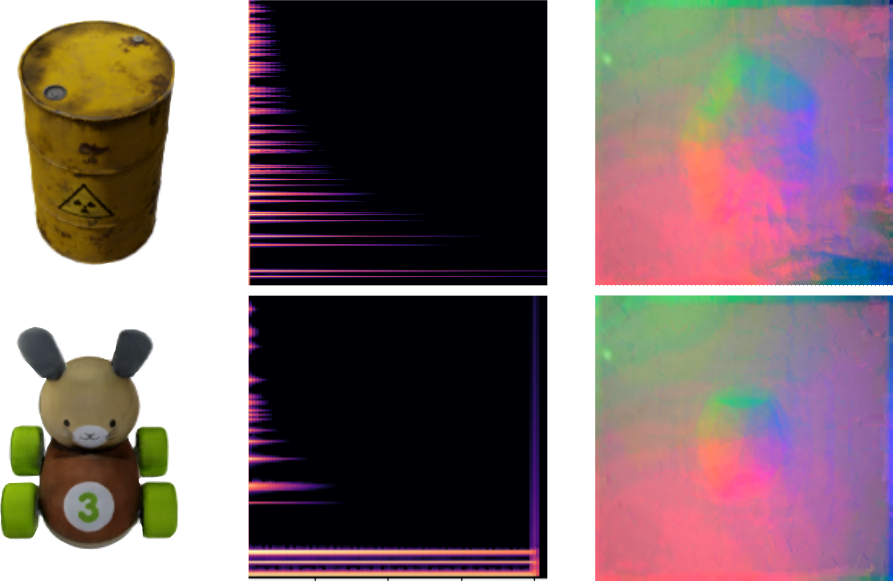}
   \caption{Visualisation of two object examples from our generated synthetic dataset. \textbf{From left to right:} The columns represent the RGB visual images, audio spectrograms, and tactile images, respectively.}
\label{fig: examples}
\end{figure}

For the purpose of this study, we generate a total of 34,500 samples of data representing 20 randomly selected objects from the ObjectFolder 2.0~\cite{gao2022objectfolder2} dataset. The objects' visual representations are demonstrated in Fig.~\ref{fig: object}. The dataset is split into training, validation, and testing subsets, each consisting of 25,500, 4,500, and 4,500 samples, respectively. Data comprises three modalities: vision, audio, and touch, for each object, as demonstrated in Fig.~\ref{fig: examples}. To render the data, a set of arguments is required, which allows to produce a diverse range of samples for each object.

\begin{itemize}
\item Visual: To generate a visual image, an array of length 6 is required, where the first three coordinates specify the camera position, and the remaining three define the position of the light source. Therefore, each testing viewpoint is described by the array: ($camera_x$, $camera_y$, $camera_z$, $light_x$, $light_y$, $light_z$).
\item Audio: To produce an audio sample, two sets of arrays are required. The first array takes three arguments specifying the point on the object surface ($x, y, z$), where force will be applied to generate an audio response. The second array, also of the same size, determines the amount of force applied to the selected coordinate point. The force is defined by providing its three directional components: ($F_{x}$, $F_{y}$, $F_{z}$).
\item Tactile: Generation of tactile images also requires two sets of arrays. Similarly to the audio parameters, the first array of length 3 defines the coordinates of the point on the object's surface. The second array requires the specification of gel rotation angle and gel displacement in the form of ($\theta$, $\phi$, $d$). 
\end{itemize}




\section{Experiment Results}
\label{Section:5}
In this section, we conduct a series of experiments to evaluate the cross-retrieval performance of our VAT-CMR model. The goal of these experiments is three-fold: (1) To assess the performance of the VAT-CMR model in comparison to competing baseline approaches; (2) To investigate how each proposed module contributes to the cross-modal retrieval performance; (3) To analyse the impact of selecting varying dominant modalities for cross-entropy learning. In the experiment, we follow~\cite{gao2022objectfolder,gao2022objectfolder2} and employ the Mean Average Precision (MAP)~\cite{rasiwasia2010new} as the metric to evaluate the retrieval performance.

\subsection{Comparison with single modality-based retrieval methods}

We compare the cross-retrieval performance between the proposed VAT-CMR model and the baseline approaches in the literature, i.e., the Canonical Correlation Analysis (CCA) method~\cite{rasiwasia2010new} and the cross-retrieval model developed in ObjectFolder~\cite{gao2022objectfolder}, as presented in Table~\ref{tab: comparison1}. To ensure an equitable comparison with the model utilised in ObjectFolder~\cite{gao2022objectfolder}, we have extended their model to accommodate multi-modal retrieval modalities. Our VAT-CMR model showcases a notable advancement over these baseline approaches when multiple retrieval modalities are used, i.e., with an increase of 0.29, 0.10 and 0.12 in MAP when employing vision, touch, and audio as the query modality, respectively. Further, our VAT-CMR consistently surpasses the baseline methods, also assessed against single-modal retrieval metrics. On the other hand, we observe that the fused representations retrieval space attains the highest MAP score, except when touch is utilised as the query modality. These results demonstrate the efficacy and versatility of our VAT-CMR model in facilitating cross-modal retrieval tasks, offering superior performance across a range of modalities and retrieval scenarios.


\begin{table}[t]
\caption{Evaluation results of cross-retrieval performance of our VAT-CMT method against CCA and ObjectFolder.} 
\label{tab: comparison1}
\begin{center}
\begin{tabular}{c c|ccc}

  Query & Retrieval & \multicolumn{3}{c}{MAP Score} \\
        &           &  Ours & CCA~\cite{rasiwasia2010new} & ObjectFolder~\cite{gao2022objectfolder}\\
  \hline
  \multicolumn{5}{c}{ 20 objects dataset (chance = 0.05)} \\  
  \hline
  Vision & Touch  & 0.93 & 0.44 & 0.51 \\
         & Audio  & 0.90 & 0.54 & 0.51 \\
         & Touch + Audio & \textbf{0.96} & --- & 0.67\\
  \hline
  Touch & Vision   &  \textbf{0.87} & 0.46 & 0.58 \\
        & Audio    &  0.84 & 0.33 & 0.73 \\
        & Vision + Audio & 0.85 & --- & 0.75\\
  \hline
  Audio  & Vision  &  0.77 & 0.37 &   0.52 \\
         & Touch   &  0.77 &  0.51 &   0.63 \\
         & Vision + Touch & \textbf{0.81} & ---& 0.69\\
  \hline
\end{tabular}
\end{center}
\end{table}

\begin{table}[t]
	\centering

		\caption{Ablation study on various network structures. }
		\label{ablation1}
            
		\begin{tabular}{cc|c c c}
                        & &  \multicolumn{3}{c}{MAP Score}  \\
            			\cline{1-5}
			 \multicolumn{2}{c|}{Network Structure} & \multicolumn{3}{c}{Query Modality} \\
            Attention & Dominant Modality  & Audio & Vision  & Touch  \\	
                        			\cline{1-5}

		&   &   0.69 & 0.67 & 0.75 \\
		&   \checkmark&    0.77 & \textbf{0.96} & 0.78 \\
        \checkmark &   &    0.76 & 0.68 & 0.81 \\
		\checkmark & \checkmark  &  \textbf{0.81} & \textbf{0.96} & \textbf{0.85} \\
		\hline
		\end{tabular}
\end{table}

\subsection{Ablation study}

In this subsection, we analyse the effects of integrating attention-based fusion and dominant modality selection into our training methodology. Specifically, when the attention module is not applied, it will be substituted with a simple feature concatenation. Meanwhile, the dominant modality selection is replaced with joint embedding-based optimisation when not in use. 

As shown in Table~\ref{ablation1}, our findings reveal that the VAT-CMR model attains the highest retrieval scores when it incorporates both dominant modality selection and the attention module within its training framework. This highlights its enhanced effectiveness in object retrieval over models lacking these components, where their absence results in an average MAP score reduction of 0.17 across all test scenarios.

Furthermore, in Table~\ref{tab: ablation2}, we provide a detailed examination of the impact of dominant modality selection on cross-modal retrieval outcomes. 
Compared to the traditional approach, our dominant modality selection method achieves an average retrieval improvement of 0.13 in the MAP score. 
Moreover, we find that there is a fixed relationship between a single modality, whether used for query or retrieval, and the dominant modality. Specifically, when audio is a single modality used for query or retrieval, selecting audio as the dominant modality in the cross-entropy learning process results in the largest average scores of 0.81 and 0.94. However having touch or vision as the dominant modality results in extremely poor average scores, as low as 0.02. This poor performance likely stems from the unique nature of audio sample representations compared to those of touch and vision, which are both represented as RGB images. 
It is also noticed that when vision is used as a single query or retrieval modality, selecting touch as the dominant modality in the cross-entropy learning process yields the largest average score. Conversely, when touch is used as a single query or retrieval modality, selecting vision as the dominant modality will obtain the largest average score. These cases also verify that touch is very similar to vision in this proposed dataset, against audio.


\begin{table}[t]
\caption{Ablation study on the selection of different dominant modalities.} 
\label{tab: ablation2}
\begin{center}
\scalebox{0.95}{
\begin{tabular}{c c| c c c | c}

  Query & Retrieval & \multicolumn{3}{c|}{Dominant Modality} & Joint\\
        &           &   Audio & Vision & Touch & \\
  \hline
  \multicolumn{6}{c}{ 20 objects dataset (chance = 0.05)} \\  
  \hline
   Audio      & Vision+Touch & \textbf{0.81} & 0.04 & 0.02 & 0.76\\
    \hline
    
    Touch    & Vision+Audio & 0.82 & \textbf{0.85} & 0.82 & 0.81\\
  \hline
    
    Vision     & Touch+Audio & 0.90 & 0.86 & \textbf{0.96} & 0.68 \\
  \hline

   Vision+Touch    &  Audio  & \textbf{0.94} & 0.06 & 0.08 & 0.81\\
  \hline
  
      Vision+Audio   & Touch  & 0.77 & \textbf{0.93} & 0.90 & 0.69\\
  \hline
  
    Touch+Audio     & Vision  & 0.78 & 0.82 & \textbf{0.85} & 0.71 \\
  \hline


\end{tabular}}
\end{center}
\end{table}




\subsection{Feature visualisation and learning curve} 
\begin{figure*}
    \centering
    \includegraphics[trim=0 0 0 -15, clip, width=\linewidth]{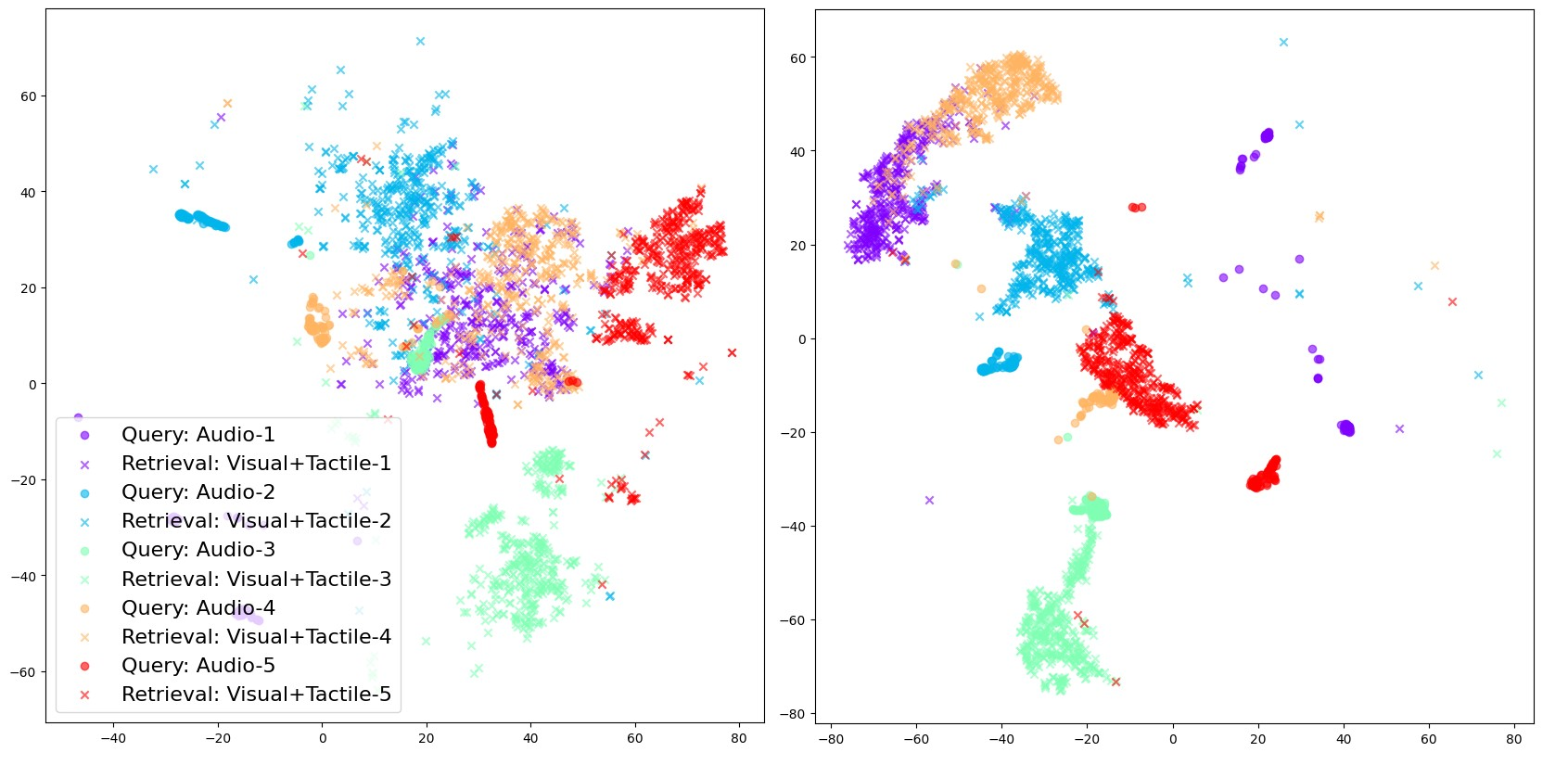}
    \caption{2D plots illustrating latent representations for a selected set of five classes, using Audio as the query test samples, and Visual+Tactile as the retrieval space. \textbf{Left:} Features after undergoing the cross-entropy model stage; \textbf{Right:} The same set of features after triplet loss processing. }
    \label{fig: t-SNE}
\end{figure*}

\begin{figure}
    \centering

    \includegraphics[width=\linewidth]{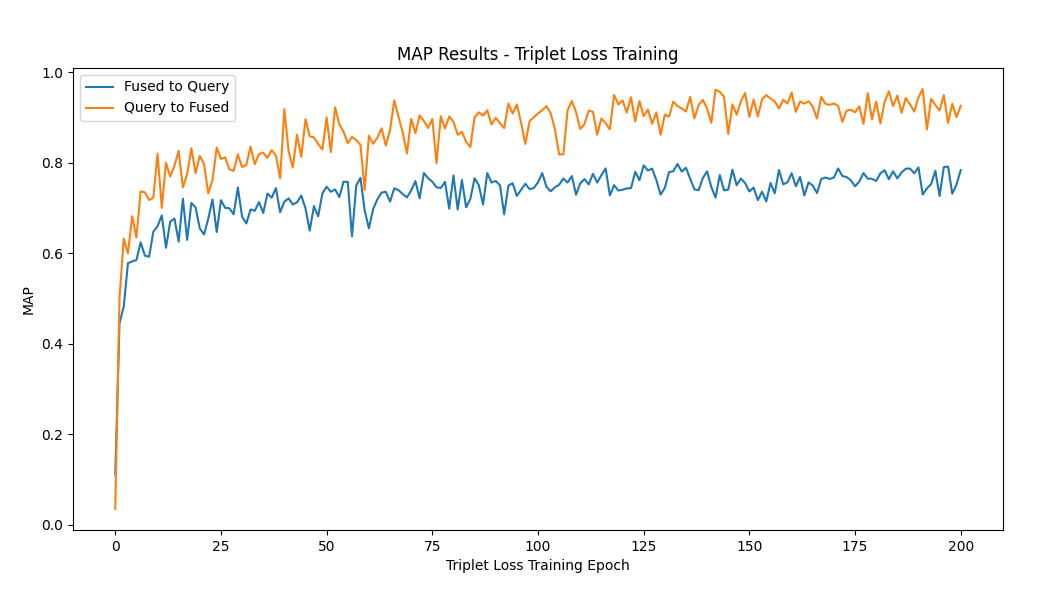}  

    \caption{Mean Average Precision evaluated on the validation dataset during cross-modal triplet loss learning with audio as the Query modality, and vision and touch as the Fused modalities.}

    \label{fig: training plot}
\end{figure}

In this subsection, we visualise the distributions of latent features obtained from our VAT-CMR model using t-SNE~\cite{van2008visualizing}. Specifically, Fig.~\ref{fig: t-SNE} shows the cross-retrieval latent space at different training phases, first after the cross-entropy loss stage and then after applying the triplet loss. The left-hand side of the figure clearly illustrates that when using only cross-entropy loss, features corresponding to the same object classes tend to cluster together, however still in a scattered manner. However, the introduction of the triplet loss stage substantially enhances feature organisation, both for the test set, and the retrieval space, leading to a more coherent clustering of representations. Furthermore, we present the MAP evaluation on the validation dataset during the triplet loss cross-modal correction learning in Fig.~\ref{fig: training plot}, further attesting to the effectiveness of the employed method.

\section{Conclusion and Discussion}
In conclusion, we have introduced VAT-CMR, a novel cross-modal retrieval model integrating vision, audio and touch modalities. When compared to previous methods, our VAT-CMR utilises a fused multi-modal retrieval representation through a multi-head attention mechanism, along with leveraging the dominant modality for cross-entropy training. Extensive experiments show that our VAT-CMR outperforms the baseline ObjectFolder method, achieving a MAP score improvement of 0.29, 0.10 and 0.12 when employing vision, touch, and audio as the query modality, respectively. A detailed ablation study further demonstrates that all the proposed modules contribute positively to cross-retrieval accuracy.


In the future work, we would like to explore a few directions based on the current findings. Firstly, in this work synthetic datasets are used, which may limit the model's generalisation to real-world scenarios, and we will design strategies to bridge the Sim2Real gap~\cite{jianu2022reducing,jing2023unsupervised}. Secondly, instead of using static visual and tactile data, we would like to consider active exploration strategies to augment robotic object retrieval capabilities so as to improve the real-world applicability of our proposed methods~\cite{cao2020spatio}. Thirdly, we will extend the study to incorporate additional perception modalities, such as force feedback, and explore zero-shot cross-modal retrieval methods~\cite{cao2024multimodal} in the context of robotic grasping. By integrating multiple modalities, we aim to enhance each sensing modality and improve overall grasping performance.

\label{Section:6}

\bibliographystyle{ieeetr}
\bibliography{egibib.bib}

\begin{thebibliography}{10}

\bibitem{kaur2021comparative}
P.~Kaur, H.~S. Pannu, and A.~K. Malhi, ``Comparative analysis on cross-modal
  information retrieval: A review,'' {\em Computer Science Review}, vol.~39,
  p.~100336, 2021.

\bibitem{luo2018vitac}
S.~Luo, W.~Yuan, E.~Adelson, A.~G. Cohn, and R.~Fuentes, ``Vitac: Feature
  sharing between vision and tactile sensing for cloth texture recognition,''
  in {\em 2018 IEEE International Conference on Robotics and Automation
  (ICRA)}, pp.~2722--2727, 2018.

\bibitem{luo2017robotic}
S.~Luo, J.~Bimbo, R.~Dahiya, and H.~Liu, ``Robotic tactile perception of object
  properties: A review,'' {\em Mechatronics}, vol.~48, pp.~54--67, 2017.

\bibitem{lee2019touching}
J.-T. Lee, D.~Bollegala, and S.~Luo, ``{“Touching to see” and “seeing to
  feel”: Robotic cross-modal sensory data generation for visual-tactile
  perception},'' in {\em 2019 International Conference on Robotics and
  Automation (ICRA)}, pp.~4276--4282, IEEE, 2019.

\bibitem{10160373}
G.~Cao, J.~Jiang, N.~Mao, D.~Bollegala, M.~Li, and S.~Luo, ``{Vis2Hap:
  Vision-based Haptic Rendering by Cross-modal Generation},'' in {\em 2023 IEEE
  International Conference on Robotics and Automation (ICRA)},
  pp.~12443--12449, 2023.

\bibitem{luo2015localizing}
S.~Luo, W.~Mou, K.~Althoefer, and H.~Liu, ``Localizing the object contact
  through matching tactile features with visual map,'' in {\em Proc. IEEE Int.
  Conf. Robot. Autom.}, pp.~3903--3908, 2015.

\bibitem{pecyna2022visual}
L.~Pecyna, S.~Dong, and S.~Luo, ``Visual-tactile multimodality for following
  deformable linear objects using reinforcement learning,'' in {\em IEEE/RSJ
  International Conference on Intelligent Robots and Systems (IROS)},
  pp.~3987--3994, 2022.

\bibitem{jiang2022shall}
J.~Jiang, G.~Cao, A.~Butterworth, T.-T. Do, and S.~Luo, ``Where shall i touch?
  vision-guided tactile poking for transparent object grasping,'' {\em
  IEEE/ASME Transactions on Mechatronics}, vol.~28, no.~1, pp.~233--244, 2022.

\bibitem{gao2022objectfolder}
R.~Gao, Y.-Y. Chang, S.~Mall, L.~Fei-Fei, and J.~Wu, ``Objectfolder: A dataset
  of objects with implicit visual, auditory, and tactile representations,'' in
  {\em Conference on Robot Learning}, pp.~466--476, PMLR, 2022.

\bibitem{ghazanfar2006neocortex}
A.~A. Ghazanfar and C.~E. Schroeder, ``Is neocortex essentially
  multisensory?,'' {\em Trends in cognitive sciences}, vol.~10, no.~6,
  pp.~278--285, 2006.

\bibitem{stein2008multisensory}
B.~E. Stein and T.~R. Stanford, ``Multisensory integration: current issues from
  the perspective of the single neuron,'' {\em Nature reviews neuroscience},
  vol.~9, no.~4, pp.~255--266, 2008.

\bibitem{rasiwasia2010new}
N.~Rasiwasia, J.~Costa~Pereira, E.~Coviello, G.~Doyle, G.~R. Lanckriet,
  R.~Levy, and N.~Vasconcelos, ``A new approach to cross-modal multimedia
  retrieval,'' in {\em Proceedings of the 18th ACM international conference on
  Multimedia}, pp.~251--260, 2010.

\bibitem{rosipal2005overview}
R.~Rosipal and N.~Kr{\"a}mer, ``Overview and recent advances in partial least
  squares,'' in {\em International Statistical and Optimization Perspectives
  Workshop" Subspace, Latent Structure and Feature Selection"}, pp.~34--51,
  Springer, 2005.

\bibitem{sharma2012generalized}
A.~Sharma, A.~Kumar, H.~Daume, and D.~W. Jacobs, ``Generalized multiview
  analysis: A discriminative latent space,'' in {\em 2012 IEEE conference on
  computer vision and pattern recognition}, pp.~2160--2167, IEEE, 2012.

\bibitem{srivastava2012multimodal}
N.~Srivastava and R.~R. Salakhutdinov, ``Multimodal learning with deep
  boltzmann machines,'' {\em Advances in neural information processing
  systems}, vol.~25, 2012.

\bibitem{andrew2013deep}
G.~Andrew, R.~Arora, J.~Bilmes, and K.~Livescu, ``Deep canonical correlation
  analysis,'' in {\em International conference on machine learning},
  pp.~1247--1255, PMLR, 2013.

\bibitem{wang2015deep}
W.~Wang, R.~Arora, K.~Livescu, and J.~Bilmes, ``On deep multi-view
  representation learning,'' in {\em International conference on machine
  learning}, pp.~1083--1092, PMLR, 2015.

\bibitem{wang2017adversarial}
B.~Wang, Y.~Yang, X.~Xu, A.~Hanjalic, and H.~T. Shen, ``Adversarial cross-modal
  retrieval,'' in {\em Proceedings of the 25th ACM international conference on
  Multimedia}, pp.~154--162, 2017.

\bibitem{zhen2019deep}
L.~Zhen, P.~Hu, X.~Wang, and D.~Peng, ``Deep supervised cross-modal
  retrieval,'' in {\em Proceedings of the IEEE/CVF Conference on Computer
  Vision and Pattern Recognition}, pp.~10394--10403, 2019.

\bibitem{liu2018active}
H.~Liu, F.~Wang, F.~Sun, and X.~Zhang, ``Active visual-tactile cross-modal
  matching,'' {\em IEEE Transactions on Cognitive and Developmental Systems},
  vol.~11, no.~2, pp.~176--187, 2018.

\bibitem{zheng2019cross}
W.~Zheng, H.~Liu, B.~Wang, and F.~Sun, ``Cross-modal surface material retrieval
  using discriminant adversarial learning,'' {\em IEEE transactions on
  industrial informatics}, vol.~15, no.~9, pp.~4978--4987, 2019.

\bibitem{liu2018surface}
H.~Liu, F.~Wang, F.~Sun, and B.~Fang, ``Surface material retrieval using weakly
  paired cross-modal learning,'' {\em IEEE Transactions on Automation Science
  and Engineering}, vol.~16, no.~2, pp.~781--791, 2018.

\bibitem{yuan2017gelsight}
W.~Yuan, S.~Dong, and E.~H. Adelson, ``Gelsight: High-resolution robot tactile
  sensors for estimating geometry and force,'' {\em Sensors}, vol.~17, no.~12,
  p.~2762, 2017.

\bibitem{lambeta2020digit}
M.~Lambeta, P.-W. Chou, S.~Tian, B.~Yang, B.~Maloon, V.~R. Most, D.~Stroud,
  R.~Santos, A.~Byagowi, G.~Kammerer, {\em et~al.}, ``Digit: A novel design for
  a low-cost compact high-resolution tactile sensor with application to in-hand
  manipulation,'' {\em IEEE Robotics and Automation Letters}, vol.~5, no.~3,
  pp.~3838--3845, 2020.

\bibitem{gomes2020geltip}
D.~F. Gomes, Z.~Lin, and S.~Luo, ``Geltip: A finger-shaped optical tactile
  sensor for robotic manipulation,'' in {\em Proc. IEEE/RSJ Int. Conf. Intell.
  Robots Syst.}, pp.~9903--9909, 2020.

\bibitem{cao2023touchroller}
G.~Cao, J.~Jiang, C.~Lu, D.~F. Gomes, and S.~Luo, ``Touchroller: A rolling
  optical tactile sensor for rapid assessment of textures for large surface
  areas,'' {\em Sensors}, vol.~23, no.~5, p.~2661, 2023.

\bibitem{he2016deep}
K.~He, X.~Zhang, S.~Ren, and J.~Sun, ``Deep residual learning for image
  recognition,'' in {\em Proceedings of the IEEE conference on computer vision
  and pattern recognition}, pp.~770--778, 2016.

\bibitem{vaswani2017attention}
A.~Vaswani, N.~Shazeer, N.~Parmar, J.~Uszkoreit, L.~Jones, A.~N. Gomez,
  {\L}.~Kaiser, and I.~Polosukhin, ``Attention is all you need,'' {\em Advances
  in neural information processing systems}, vol.~30, 2017.

\bibitem{zheng2020lifelong}
W.~Zheng, H.~Liu, and F.~Sun, ``Lifelong visual-tactile cross-modal learning
  for robotic material perception,'' {\em IEEE transactions on neural networks
  and learning systems}, vol.~32, no.~3, pp.~1192--1203, 2020.

\bibitem{gao2022objectfolder2}
R.~Gao, Z.~Si, Y.-Y. Chang, S.~Clarke, J.~Bohg, L.~Fei-Fei, W.~Yuan, and J.~Wu,
  ``Objectfolder 2.0: A multisensory object dataset for sim2real transfer,'' in
  {\em Proceedings of the IEEE/CVF conference on computer vision and pattern
  recognition}, pp.~10598--10608, 2022.

\bibitem{van2008visualizing}
L.~Van~der Maaten and G.~Hinton, ``Visualizing data using t-sne.,'' {\em
  Journal of machine learning research}, vol.~9, no.~11, 2008.

\bibitem{jianu2022reducing}
T.~Jianu, D.~F. Gomes, and S.~Luo, ``Reducing tactile sim2real domain gaps via
  deep texture generation networks,'' in {\em 2022 International Conference on
  Robotics and Automation (ICRA)}, pp.~8305--8311, IEEE, 2022.

\bibitem{jing2023unsupervised}
X.~Jing, K.~Qian, T.~Jianu, and S.~Luo, ``Unsupervised adversarial domain
  adaptation for sim-to-real transfer of tactile images,'' {\em IEEE
  Transactions on Instrumentation and Measurement}, vol.~72, pp.~1--11, 2023.

\bibitem{cao2020spatio}
G.~Cao, Y.~Zhou, D.~Bollegala, and S.~Luo, ``Spatio-temporal attention model
  for tactile texture recognition,'' in {\em Proc. IEEE/RSJ Int. Conf. Intell.
  Robots Syst.}, pp.~9896--9902, 2020.

\bibitem{cao2024multimodal}
G.~Cao, J.~Jiang, D.~Bollegala, M.~Li, and S.~Luo, ``Multimodal zero-shot
  learning for tactile texture recognition,'' {\em Robotics and Autonomous
  Systems}, vol.~176, p.~104688, 2024.

\end{thebibliography}

\end{document}